\documentclass[10pt, letter,twocolumn]{IEEEtran}
\usepackage{amsmath} 
\usepackage{amsmath}
\usepackage{gensymb}
\usepackage{balance}
\usepackage{float}
\usepackage{romannum}
\usepackage{graphicx}
\usepackage{comment}
\usepackage{subcaption}
\usepackage{placeins}
\usepackage{algorithm}
\usepackage{algorithmic}

\usepackage{amsmath}
\usepackage{amssymb}
\usepackage{amsxtra}
\usepackage{multirow}
\usepackage{mathtools}
\usepackage{cite}	
\usepackage{color}
\usepackage{bbm}
\usepackage[numbers]{natbib}

\title{Leveraging 3D LiDAR Sensors to Enable Enhanced Urban Safety and Public Health: Pedestrian Monitoring and Abnormal Activity Detection\vspace{-0.2cm}} 
\author{\IEEEauthorblockN{\large Nawfal Guefrachi$^{1}$, Jian Shi$^{2}$, Hakim Ghazzai$^{2}$, and Ahmad Alsharoa$^{1}$}\\\IEEEauthorblockA{\small $^{1}$ Missouri University of Science and Technology(MST), Rolla, Missouri, USA\\
$^{2}$CEMSE Division, King Abdullah University of Science and Technology (KAUST), Thuwal, Saudi Arabia
}
\vspace{-0.5cm}}
\begin{document}
\maketitle
\thispagestyle{empty}
\begin{abstract}
\boldmath
The integration of Light Detection and Ranging (LiDAR) and Internet of Things (IoT) technologies offers transformative opportunities for public health informatics in urban safety and pedestrian well-being. This paper proposes a novel framework utilizing these technologies for enhanced 3D object detection and activity classification in urban traffic scenarios. By employing elevated LiDAR, we obtain detailed 3D point cloud data, enabling precise pedestrian activity monitoring. To overcome urban data scarcity, we create a specialized dataset through simulated traffic environments in Blender, facilitating targeted model training. Our approach employs a modified Point Voxel-Region-based Convolutional Neural Network (PV-RCNN) for robust 3D detection and PointNet for classifying pedestrian activities, significantly benefiting urban traffic management and public health by offering insights into pedestrian behavior and promoting safer urban environments. Our dual-model approach not only enhances urban traffic management but also contributes significantly to public health by providing insights into pedestrian behavior and promoting safer urban environment.
\end{abstract}
\vspace{-0.15cm}
\vspace{-0.4cm}
\section{Introduction}
The ability to accurately detect and classify objects in three-dimensional (3D) space is increasingly crucial for traffic management systems, especially in the context of public health monitoring within urban environments. Advanced 3D object detection technologies, including Light Detection and Ranging (LiDAR), play a pivotal role in ensuring pedestrian safety by providing precise data on pedestrian movements and behaviors. The integration of Internet of Things (IoT) technologies further enhances these capabilities, offering promising avenues to improve public safety and health outcomes by facilitating more efficient pedestrian monitoring and management~\cite{ahmed2022smart}.
Prior advancements in pedestrian monitoring have predominantly utilized camera-based systems, as highlighted in \cite{gawande2023real}. Their work presents a deep learning approach for pedestrian detection and suspicious activity recognition. This study underscores the significance of video surveillance in enhancing security through real-time tracking and behavior analysis in various settings. Despite its innovations, this camera-based method, like others before it, encounters challenges such as limited environmental adaptability and privacy concerns, which underscore the necessity for alternative technologies and the imperative for advanced and innovative solutions.

In \cite{6347144}, the authors propose enhancing pedestrian safety using body-mounted depth cameras. While innovative, this approach suffers from limitations like a restricted field of view and potential obstructions in crowded areas. In contrast, our research utilizes elevated LiDAR technology, providing comprehensive spatial awareness and overcoming these issues, resulting in a more reliable pedestrian safety system. While wearable sensors show potential, as explored in \cite{8662658}, they face limitations such as user discomfort and challenges in dynamic monitoring. Our focus is on 3D LiDAR technology, a non-intrusive solution that integrates into urban infrastructure, enabling detailed behavioral analysis and enhancing urban planning and safety strategies. LiDAR technology, with laser-based 3D mapping, excels in tracking pedestrian movements across diverse conditions, offering high precision and data-driven insights while safeguarding privacy \cite{rinchi2023lidar}. In contrast to prior research using datasets like KITTI \cite{geiger2013vision} for autonomous applications, our work advances urban pedestrian monitoring with a specialized dataset, enhancing accuracy and context relevance in pedestrian safety technologies. While 3D LiDAR has been used for pedestrian monitoring, it previously focused only on human detection. Building upon the foundational work in \cite{WANG201771}, which introduced pedestrian detection using 3D LiDAR and SVM classifiers, our research innovates by not only detecting but also classifying pedestrian activities. This innovation enhances the insights provided in \cite{zhao2019detection, app12041799}, emphasizing a comprehensive understanding of pedestrian activities, which is pivotal for the development of safety and interaction protocols in autonomous systems. The focus on precise activity classification underscores a significant advance in urban mobility and pedestrian monitoring technology, contributing to unexplored avenues for improving pedestrian health and safety and enriching the integration of advanced pedestrian surveillance technologies into urban systems.

{In our work, we strategically deploy elevated LiDAR sensors across urban infrastructures to conduct extensive pedestrian monitoring. This methodology is aimed at accurately detecting a range of pedestrian activities, both typical and atypical, to directly tackle public health and pedestrian safety issues. Initiating with the collection of detailed 3D point cloud data, our approach provides unmatched depth in understanding pedestrian behaviors within densely populated urban environments. By effectively distinguishing between normal and abnormal pedestrian activities, we empower the development of precise public health interventions designed to significantly improve pedestrian safety and health. This proactive monitoring framework is crucial for pinpointing and mitigating potential health risks, underscoring our contribution to the field of public health informatics with emphasis on enhancing pedestrian well-being.}
\begin{table}[t!]
\centering
\caption{Pedestrian activity classes}
\renewcommand{\arraystretch}{1.1}
\begin{tabular}{|p{3.6cm}|p{3.6cm}|}
\hline
\textbf{Normal Behavior}  & \textbf{Abnormal Behavior}\\
\hline
Walking & Dizzy walking\\
 \hline
Running  &Falling\\
 \hline
Talking on the phone & Walking with injured leg\\ 
\hline
\end{tabular}
\label{tab:Activity Classes}
\end{table}
\vspace{-12mm}
\section{Proposed Methodology}
\begin{figure}[t!]
\centering
 \includegraphics[width=8.6cm,height=5.5cm]{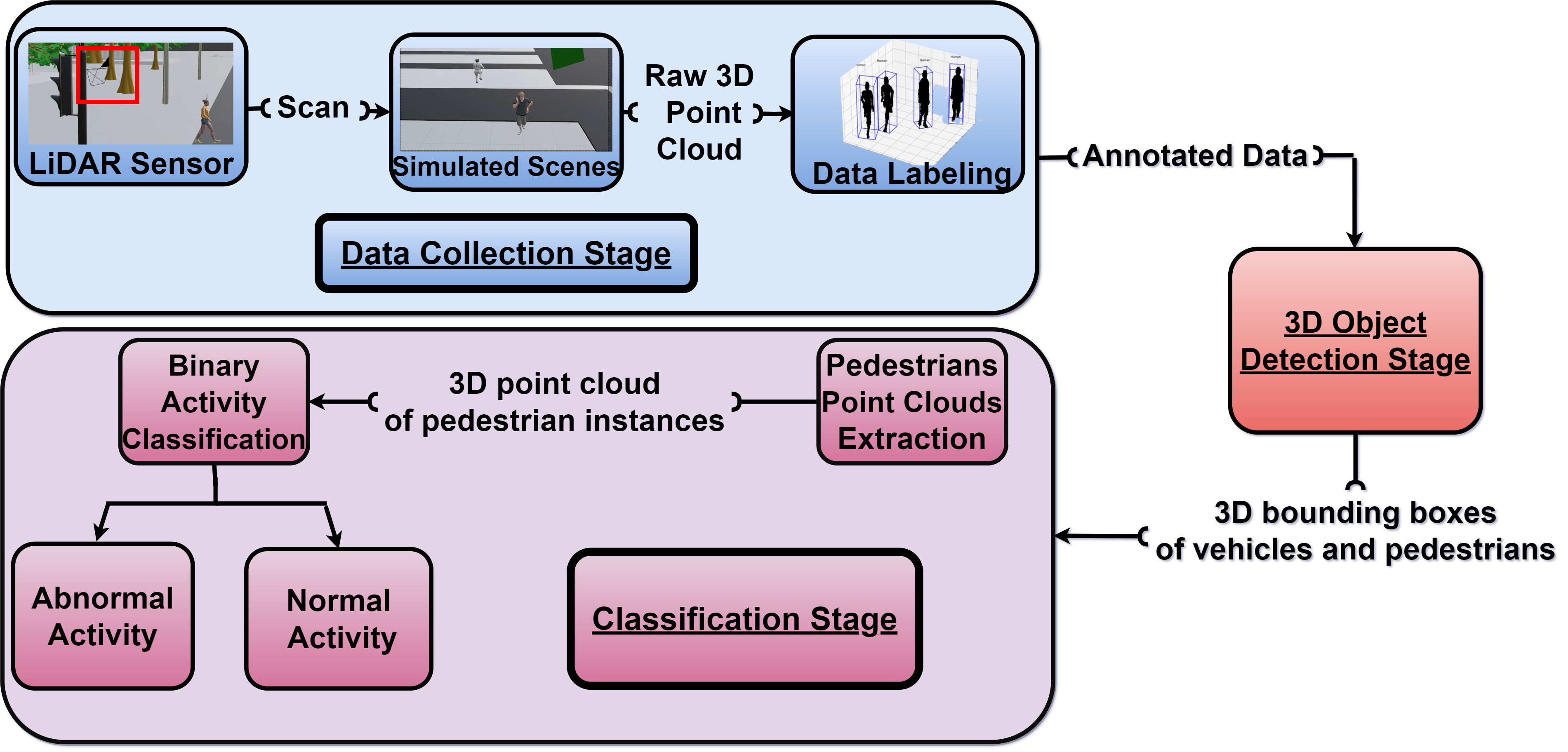}\vspace{-1mm}
\caption{Workflow of our proposed framework for pedestrian detection and activity classification.} 
\vspace{-1mm}
\label{fig1}
\end{figure}

Our main objective is to create a computer vision framework that uses LiDAR point cloud data to detect 3D objects. We aim to output 3D bounding boxes for specific objects of interest, namely vehicles and pedestrians. Additionally, we classify the activities of the detected pedestrian instances. As shown in Figure~\ref{fig1}, we propose a method that includes three main phases: Phase 1 involves data collection and annotation; Phase 2 focuses on the use of 3D object detection to generate 3D bounding boxes; Phase 3 centers on extracting 3D point clouds related to pedestrians and classifying their activities. To accomplish our goals, we follow these steps:
\begin{itemize}
 \item \textbf{Data Collection and Labeling}: We collect LiDAR data to capture 3D aspects of traffic situations, serving as input for our computer vision framework. This data is meticulously labeled, especially for vehicles and pedestrians, involving entity identification and marking relevant points within the point clouds.
 \item \textbf{3D Object Detection}: We refine a deep learning model to process labeled data effectively, identifying both vehicles and pedestrians as moving objects and pinpointing their 3D positions. By focusing our analysis on these two dynamic categories, we significantly minimize the risk of misclassification, thereby enhancing the precision of our pedestrian activity classifications. This tailored approach ensures a more accurate determination of each entity's location, crucial for understanding urban mobility patterns and improving pedestrian safety measures.
\item \textbf{Pedestrian Activity Classification}: In this stage, we process point clouds from identified pedestrians for input into our classification model. Our framework effectively utilizes these datasets to classify pedestrian activities as 'Normal' or 'Abnormal'. 'Normal' behavior signifies typical, expected movements, while 'Abnormal' behavior may indicate health concerns, such as injury, dizyness or distress, necessitating further investigation or intervention as shown in Table ~\ref{tab:Activity Classes}.
\end{itemize}
Importantly, our innovative framework revolutionizes pedestrian monitoring by strategically positioning elevated LiDAR sensors on various urban infrastructures, including traffic lights and street lamps. This positioning allows us to capture detailed 3D point cloud data of pedestrian movements below, providing comprehensive coverage, accuracy, and reliability. This high-quality data enhances pedestrian detection and activity classification, ultimately improving the effectiveness of urban monitoring systems.
\vspace{-2mm}
\section{Elevated LiDAR-based Pedestrian Monitoring}
\subsection{Data Collection}
\begin{table}[t!]
\centering
\caption{3D point cloud format}
\begin{tabular}{ |p{3.8cm}|p{3.8cm}| }
\hline
\textbf{Parameters} & \textbf{Definition} \\
\hline
(x, y, z) & 3D coordinates of the object center \\
\hline
(W, H, L) & Width, Height, Length of the bounding box \\
\hline
Heading\textunderscore angle & Rotation angle according to the Z-axis within the scene world \\
\hline
Category\textunderscore name & Class name of the object of interest \\
\hline
Activity\textunderscore Classification (Pedestrians Only) & Class of the Pedestrian Behavior \\
\hline
\end{tabular}
\label{tab:BB}
\vspace{-3mm}
\end{table}
\subsubsection{Collection of Raw Point Cloud Data}
\begin{figure}[t]
  \centering
  \includegraphics[width=7.5cm, height=5.5cm]{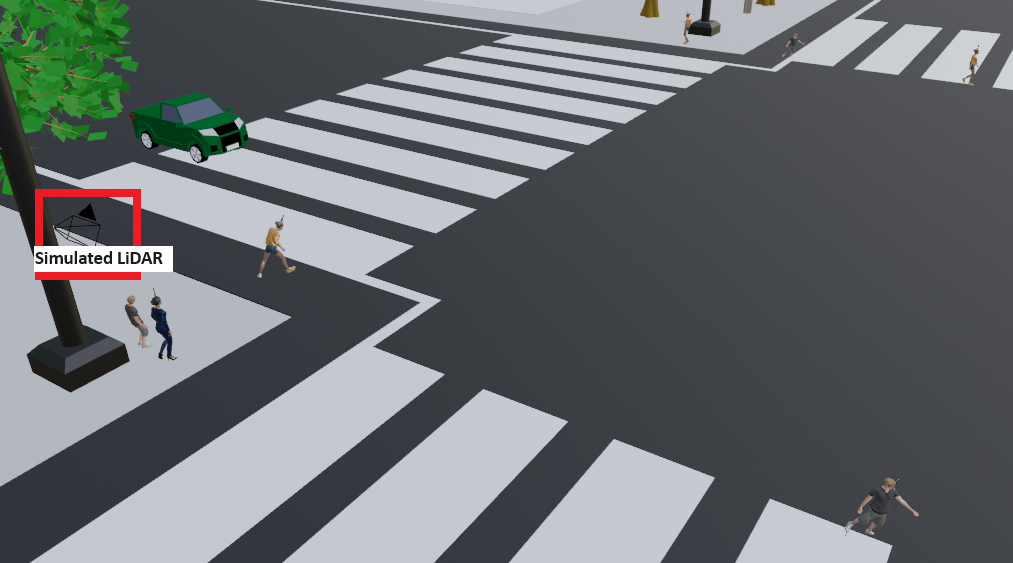}
  \caption{Screenshot of a simulated scene from Blender with an elevated LiDAR.}
  \vspace{-6mm} 
  \label{Scene}
\end{figure}

Our study focuses primarily on pedestrian monitoring, which necessitates real-world datasets obtained from elevated LiDAR systems capturing diverse scenarios. However, collecting such data faces challenges, particularly in creating potentially dangerous scenarios for research purposes. Safety concerns, especially for risky activities like falls, prevent intentional scenario creation. As an alternative, we propose using a simulated environment to accurately replicate the required scenarios. For simulating traffic scenes with moving vehicles and pedestrians, and collecting data from a simulated LiDAR mounted on a traffic light, we employ Blender \cite{reitmann2021blainder}. Blender, recently adopted for 3D point cloud data collection and editing, offers a wide range of features, including animation, object manipulation, and a LiDAR simulation add-on essential for authentic LiDAR data collection. This add-on accurately represents reflected beam intensity values for different materials. Using Blender, we create various scenes depicting humans engaged in normal and abnormal activities, integrating different vehicles as shown in Figure ~\ref{Scene}. Strategically positioned LiDAR sensors at a 3-meter height and angled downward capture 3D coordinates and reflectivity values of objects, enhancing our dataset for comprehensive analysis.

\subsubsection{Data Annotation}
To effectively utilize the collected raw data as input for our model, it is essential to thoroughly annotate the data points. This annotation process involves meticulously creating accurate 3D bounding boxes that precisely define the position of each object within the scene, accommodating various scenarios. The bounding boxes we develop adhere to a specific, well-defined format outlined in Table \ref{tab:BB}. This format incorporates comprehensive details such as object identifiers, coordinates and dimensions.\break Undertaking this step is absolutely crucial for ensuring proper data preparation, which in turn greatly facilitates and streamlines subsequent processes such as model training and thorough evaluation. By consistently employing a standardized format throughout the entire data annotation phase, we ensure a high level of consistency and coherence. This methodical approach significantly enhances both the accuracy and the reliability of our labeled dataset, making it a valuable resource for our research and analysis. This meticulous preparation is key to achieving meaningful and trustworthy results.
\FloatBarrier
\vspace{-3mm}
\subsection{3D Object Detection}
We use the Point Voxel-Region-based Convolutional Neural Network (PV-RCNN) as our primary 3D object detection architecture for point clouds \cite{shi2020pv}. We feed the raw 3D point cloud data directly into our system. This data, filled with attributes like 3D coordinates and intensity, goes through a transformation within our architecture. We meticulously fine-tune the PV-RCNN's parameters, which enhances its adaptability and precision. We set the maximum bounding box dimensions, with human boxes at (1.7, 1.7, 2) meters and vehicle dimensions at (3, 3, 3.5) meters. Our voxel feature encoding captures attributes such as geometric traits. We adjust the VoxelSetAbstraction parameters, selecting 4096 keypoints using the Furthest Point Sampling (FPS) method. For refining features, we employ the backbone network with the Adam OneCycle optimizer, an initial learning rate of 0.01, and a momentum of 0.9. In our RoI generation and pooling stages, we set rotation constraints between 0 and 1.57 radians, ensuring accurate object localization within the point clouds. At its core, our PV-RCNN integrates both voxel and point cloud methodologies seamlessly. This balance lets us dive deep into the 3D data's details and capture the broader context within the scene. By incorporating these fine-tuned parameters, our PV-RCNN becomes a powerful object detection framework, adapting flexibly to various applications. This blend of architecture and application-specific details ensures our 3D object detection is both detailed and comprehensive. 
\vspace{-4mm}
\subsection{Pedestrian Activity Classification}
Following the detection phase, the subsequent step involves the collection of 3D point cloud data linked with identified pedestrian instances for classification. This extraction technique aims to pinpoint and concentrate on specific point clouds relevant to the observed pedestrians. During the training phase, these points are identified by using a predetermined Intersection over Union (IoU) threshold to compare actual data with the predicted bounding boxes. If the overlap surpasses the set threshold of 0.65, the system applies the actual data's activity label to the recognized entity. Once pedestrian-related points are sorted, each bounding box is assigned a unique identifier, linking the 3D points and their dimensions to a particular pedestrian instance. Afterwards PointNet architecture is utilized for classification, determining if each instance is "Normal" or "Abnormal" \cite{qi2017pointnet}. Our approach entails a two-step process. PointNet is specifically designed for 3D point cloud data, showcasing its utility across various computer vision applications. By exploiting its capabilities, we tailor it for pedestrian activity classification. PointNet uniquely processes 3D point cloud data, maintaining consistency despite different transformations, highlighting its role in 3D perception tasks. It includes the Input Transform Net for initial point cloud data transformation, ensuring rotation and translation invariance. The data undergoes processing to capture local and global features, with shared multi-layer perceptrons (MLPs) and max-pooling operations forming a comprehensive global feature vector. The Point-wise Feature Transformer enhances PointNet's performance, focusing on calculating a transformation matrix for each point, ensuring model invariance to point permutations. Afterwards, a feature propagation module is added, improving feature refinement by leveraging inter-point connections, enriching feature learning with contextual information. Following feature extraction and transformation, PointNet uses fully connected layers or MLPs to map features to outputs for final classification. We modify the architecture for binary classification, with a two-neuron output layer and softmax activation for class probability determination. Fine-tuning includes a categorical cross-entropy loss function to combat overfitting, integrating dropout layers throughout the model. Adopting PointNet illustrates the effectiveness of 3D point cloud data in pedestrian activity classification, ensuring data consistency despite variations. By capturing both detailed and broad features, PointNet excels in accurately identifying and classifying pedestrian activities, adeptly recognizing various movements and actions. This capability ensures reliable and consistent outcomes, even with data position or orientation changes.
\vspace{-5mm}
\section{Results and Discussion}
\subsection{Dataset and Evaluation Metrics}
\begin{figure}[t]
  \centering
  \includegraphics[width=7.5cm, height=5.5cm]{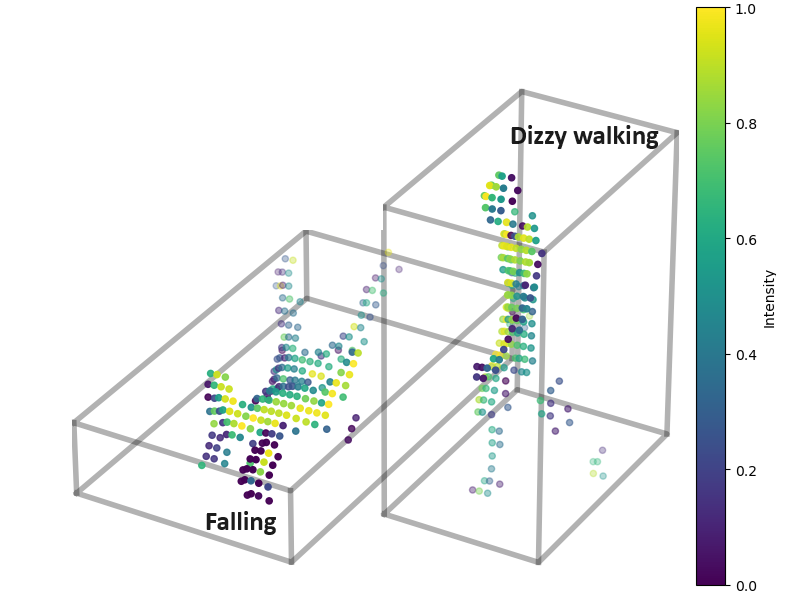}
  \caption{Dataset Visualization:  This screenshot highlights pedestrians. The color variations stem from different reflectivity values of the point cloud components.}
  \label{vis}
  \vspace{-6mm}
\end{figure}

\vspace{-1mm}
To enhance our dataset's utility for pedestrian safety and health monitoring, we meticulously crafted 21 diverse urban scenarios in Blender, each teeming with vehicles and pedestrians of all ages engaged in a broad spectrum of activities. This meticulous simulation captures the essence of pedestrian dynamics, crucial for identifying and categorizing behaviors as 'Normal' and 'Abnormal' as shown in Table ~\ref{tab:Activity Classes}. For instance, the inclusion of 'Walking' under normal behaviors versus 'Dizzy Walking' as abnormal directly supports the development of deep learning models aimed at recognizing potential safety threats or health concerns, thereby underscoring our commitment to elevating urban safety measures and public health strategies. The scenes are further enriched with intricate animations, contributing to a robust dataset comprising 550 to 2500 frames per scene. Strategically placed LiDAR sensors within each scene capture extensive 3D point clouds, amassing over 350000 point per frame, to facilitate nuanced object detection and activity analysis. A visual representation of a single frame from this dataset is shown in Figure ~\ref{vis}. 
To accurately evaluate and comprehensively assess the performance of our methodology in both 3D object detection and pedestrian activity classification, we carefully adopt  metrics tailored to each distinct task.\break Importantly, some metrics, due to their broad applicability, are leveraged across both tasks.
For 3D object detection, our focus lies on:
\begin{itemize}
    \item \textbf{Average Precision (AP)}: By summarizing the precision-recall curve over all recall values, offers a comprehensive measure for object detection.
    \item \textbf{Recall}: Emphasizing the model's ability to identify all relevant instances, it ensures the vast majority of objects are detected.
    \item \textbf{Precision}: Highlighting the accuracy of positive identifications, it is pivotal for validating the confidence in our classifications.
    \item \textbf{F1-Score}: Serving as the harmonic mean of precision and recall, it provides a balanced measure, encapsulating both detection and classification accuracy
\end{itemize}
When classifying pedestrian activities, we lean on the already mentioned metrics and for an overall insights, we also focus on the overall accuracy which is a general measure that portrays the correct predictions as a proportion of total predictions.
\vspace{-0.4cm}
\subsection{Detection Results}
Before examining the detection metrics of PV-RCNN and SECOND, it is crucial to note their architectural distinctions. PV-RCNN blends point and voxel features for detailed 3D detection, whereas SECOND emphasizes speed with a voxel-only approach. Table \ref{tab:detection} provides a comparative analysis of detection metrics for two prominent 3D object detection models: PV-RCNN and SECOND. PV-RCNN, characterized by its hybrid design combining point and voxel features, demonstrates superior performance in both pedestrian and vehicle detection. In the realm of detecting pedestrians, PV-RCNN achieves a remarkable F1-Score of 82.91\%, outperforming SECOND, which attains 74.49\%. This performance superiority extends to vehicle detection as well, with PV-RCNN consistently achieving higher metrics. Specifically, PV-RCNN attains an F1-Score of 84.32\% for vehicle detection, while SECOND achieves 77.38\%. These results underscore PV-RCNN's advanced detection capabilities, owing to its ability to effectively leverage 3D point cloud data by synergizing detailed semantic information with spatial geometry. This approach enhances the model's capacity to accurately identify pedestrians and vehicles, even in complex urban environments, contributing to reduced false positives and improved detection accuracy.
\FloatBarrier
\begin{table}[htbp]
  \centering
  \caption{Detection results comparison: PV-RCNN vs. SECOND}
  \label{tab:detection}
  \begin{tabular}{|l|p{1.3cm}|p{1.3cm}|p{1.3cm}|p{1.3cm}|}
    \hline
     \textbf{Class}& \multicolumn{2}{c|}{\textbf{Pedestrians}} & \multicolumn{2}{c|}{\textbf{Vehicles}} \\
    \hline
    \textbf{Metric} & \textbf{PV-RCNN} & \textbf{SECOND} & \textbf{PV-RCNN} & \textbf{SECOND} \\
    \hline
    AP & \textbf{83.32\%} & 74.36\% & \textbf{87.14\%} & 77.27\% \\
    \hline
    Precision & \textbf{85.53\%} & 72.23\% & \textbf{86.54\%} & 75.71\% \\
    \hline
    Recall & \textbf{87.82\%} & 76.89\% & \textbf{88.95\%} & 79.12\% \\
    \hline
    F1-Score & \textbf{82.91\%} & 74.49\% & \textbf{84.32\%} & 77.38\% \\
    \hline
  \end{tabular}
\end{table}
\vspace{-6.5mm}
\subsection{Classification Results}
\begin{figure}[t]
  \centering
  \includegraphics[width=8cm, height=7cm]{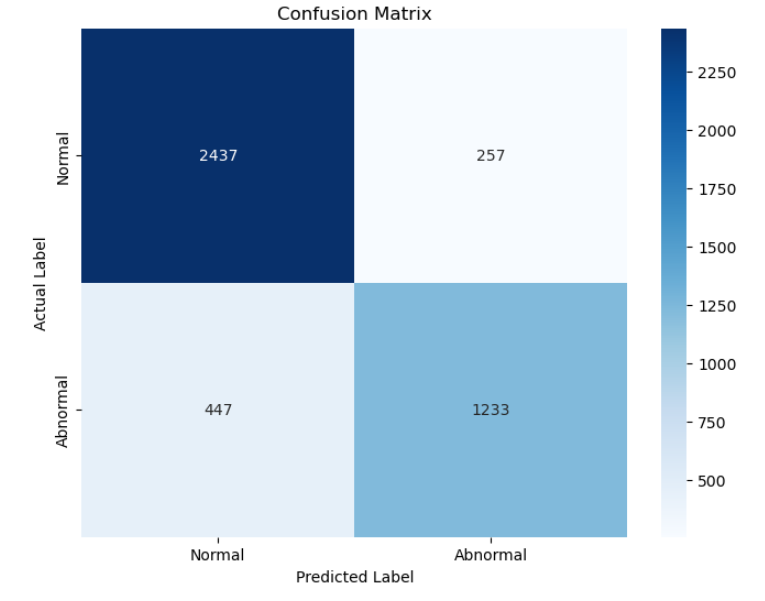}
  \caption{Confusion matrix showcasing the classification results of pedestrian activities into 'Normal' and 'Abnormal'.}
  \label{confusion}
  \vspace{-6mm}
\end{figure}

\vspace{-0.5mm}
The results obtained from the PointNet model are presented in the confusion matrix, which encompasses a total of 4374 instances, as illustrated in Figure ~\ref{confusion}. This comprehensive evaluation underscores the model's remarkable capability to effectively discern between normal and abnormal human activities. Notably, when categorizing instances as 'Normal,' the model exhibited an impressive accuracy rate, correctly identifying 2437 instances. However, it is worth noting that there were 257 instances that were misclassified as 'Abnormal,' indicating space for improvement in precision. Conversely, when tasked with classifying instances as 'Abnormal,' the PointNet model demonstrated proficiency by accurately classifying 1233 instances. Nonetheless, there were 447 instances that were falsely categorized as 'Abnormal,' signifying an area where further enhancement is warranted. In sum, the results portray a robust model, particularly in the context of 'Normal' activities. Tables \ref{tab:Models} and \ref{tab:classification} show a comparison of classification metrics between PointNet and Voxel-Based MLP for classifying normal and abnormal behaviors. We look at several important measures like Overall Accuracy, Precision, Recall, and F1-Score to evaluate performance. The classification comparison contrasts Voxel-Based MLP's grid analysis with PointNet's direct point clouds processing. Voxel-Based MLP converts data into a voxel grid for feature analysis, while PointNet's architecture allows for intricate pedestrian activity classification, showcasing its effectiveness in our framework. PointNet demonstrates a noticeably enhanced performance in comparison to Voxel-Based MLP when it comes to classifying pedestrian behaviors. Specifically, it stands out with an impressive 83.92\% accuracy rate in accurately identifying normal behaviors. This level of performance is significantly higher when set against the 67.72\% accuracy rate that Voxel-Based MLP achieves. This marked contrast in their accuracies highlights PointNet's enhanced effectiveness and precision in behavior classification tasks. PointNet achieves greater precision (84.51\% for normal and 82.74\% for abnormal behaviors) and a higher recall rate for normal behavior at 90.47\%.\break Achieving 87.40\% as F1-score for normal behavior demonstrates a balanced performance. This superior performance is due to its direct exploitation of spatial data from point clouds, outperforming the Voxel-Based MLP which potentially neglects key aspects of spatial relationships and details during the conversion of data points.
\vspace{-1mm}
\begin{table}[t!]
\centering
\caption{Overall accuracy comparison: PointNet vs. Voxel-Based MLP}
\begin{tabular}{ |p{3.6cm}|p{3.6cm}| }
\hline
\textbf{Model} & \textbf{Overall Accuracy} \\
\hline
PointNet & \textbf{83.92\%} \\
\hline
Voxel MLP & 67.72\% \\
\hline
\end{tabular}
\label{tab:Models}
\vspace{-0.2cm}
\end{table}

\begin{table}[thtbp]
  \centering
  \caption{Classification results comparison: PointNet vs. Voxel-Based MLP}
  \label{tab:classification}
  \begin{tabular}{|l|p{1.1cm}|p{1.35cm}|p{1.1cm}|p{1.35cm}|}
    \hline
    \textbf{Class} & \multicolumn{2}{c|}{\textbf{Normal behavior}} & \multicolumn{2}{c|}{\textbf{Abnormal behavior}} \\
    \hline
    \textbf{Metric}& \textbf{PointNet} & \textbf{Voxel MLP} & \textbf{PointNet} & \textbf{Voxel MLP} \\
    \hline
    Precision & \textbf{84.51\%} & 73.33\% & \textbf{82.74\%} & 64.70\% \\
    \hline
    Recall & \textbf{90.47\%} & 72.68\% &\textbf{73.38}\% & {72.68\%} \\
    \hline
    F1-Score & \textbf{87.40\%} & 73.00\% & \textbf{77.76\%} & 68.40\% \\
    \hline
  \end{tabular}
\vspace{-1mm}
\end{table}

Our work enhances pedestrian safety by applying PointNet to analyze urban pedestrian behaviors accurately. This pivotal contribution showcases the model's effectiveness in handling complex spatial data, marking a significant advance in monitoring technology and laying groundwork for improved urban safety strategies.
\vspace{-3mm}
\section{Conclusion}
In this paper, we have explored pedestrian behavior classification to boost public health using 3D LiDAR-based point cloud data within a three-part framework. Initially, we create a dataset using the Blender simulator, then apply PV-RCNN for precise pedestrian detection. Next, we extract 3D point cloud data for pedestrian instances and feed them to PointNet for binary activity classification. This method accurately differentiates 'Normal' from 'Abnormal' behaviors, offering a detailed analysis of pedestrian activities. By incorporating advanced technologies, our framework significantly improves pedestrian activity monitoring, essential for public health, by identifying behavioral patterns that proactively indicate public health risks, thus enhancing safety and well-being.
\bibliographystyle{IEEEtran}
\vspace{-1mm}
\bibliography{References}
\end{document}